\documentclass[]{fairmeta}

\usepackage{amssymb}
\usepackage{amsmath}
\usepackage{algorithm2e}
\usepackage[most]{tcolorbox}

\usepackage{graphicx}
\usepackage{wrapfig}

\newcommand{\argmin}{\operatorname*{arg\,min}}
\newcommand{\argmax}{\operatorname*{arg\,max}}
\newcommand{\Nmax}{N_{\text{max}}}
\newcommand{\Dmax}{D_{\text{max}}}
\newcommand{\tmax}{\theta_{\text{max}}^*}

\title{Efficiently Estimating Optimal Hyperparameter Scaling Laws through Power-Law Entropy Search}

\author[1,2]{Zhiliang Chen}
\author[1]{Sebastian Ament}
\author[1]{David Eriksson}
\author[3]{Maximilian Balandat}
\author[2]{Bryan Kian Hsiang Low}
\author[1]{Eytan Bakshy}
\author[1]{Jihao Andreas Lin}

\contribution[1]{Meta}
\contribution[2]{National University of Singapore}
\contribution[3]{Atomic Machines; work done at Meta}

\abstract{Optimal hyperparameter scaling laws describe how the best hyperparameters for large language model (LLM) training change with model and data scale, enabling practitioners to predict optimal configurations at production scales without expensive large-scale tuning. However, estimating these scaling laws conventionally requires exhaustive grid searches over thousands of training runs, consuming enormous computational resources. We introduce Power-Law Entropy Search (\textbf{PLES}), a computational cost-aware acquisition function built on multi-fidelity Bayesian optimization that efficiently estimates optimal hyperparameter scaling laws through adaptive experimentation. A key innovation in \textbf{PLES} is that it searches for candidates that reduce the overall uncertainty of a scaling law estimate, instead of optimizing a single objective function. At each iteration, \textbf{PLES} selects the candidate configuration that maximally reduces the uncertainty of the scaling law estimates per unit computational cost, naturally favoring informative small-scale experiments. We evaluate \textbf{PLES} on synthetic benchmarks, surrogate models fitted to real LLM training data, and actual LLM pre-training runs. Across all settings, \textbf{PLES} converges to accurate optimal hyperparameter scaling laws using less than one-tenth of the computational budget required by conventional grid search and other baselines.}

\date{\today}
\correspondence{\email{chenzhiliang@u.nus.edu, jandylin@meta.com}}

\begin{document}

\maketitle

\section{Introduction}
\label{sec:intro}

Optimal hyperparameter scaling laws \citep{li2025predictablescaleistep} describe how optimal hyperparameters (e.g., learning rate, batch size, weight decay, etc) change w.r.t. increasing model and data sizes. These scaling laws are critical tools to guide the development of large language models (LLMs) for real-world applications \citep{chen2025broadenscopeefficientmultiturn} because they allow practitioners to find optimal hyperparameters for production-level model and data scales without actually performing any hyperparameter optimization (HPO) \citep{feurer2019hyperparameter,akiba2019optuna} at those scales.

It is costly to derive these scaling laws. This is because optimal hyperparameter scaling laws requires us to identify optimal hyperparameters at every scaling ladder (which forms the subtle difference from conventional scaling laws). A typical approach to find optimal hyperparameter scaling laws is through repeated experimentation over a large grid of hyperparameters. Running these experiments is unfortunately computationally expensive. For instance, \citet{li2025predictablescaleistep} "t\textit{rained over 3,700 LLMs from
scratch across 100 trillion tokens, consuming nearly one million NVIDIA H800
GPU hours}" to establish an optimal scaling law for learning rate and batch size. In addition, scaling laws become outdated with newer model architectures and settings, requiring us to often repeat such expensive grid-search approaches.

We introduce a new adaptive experimentation method that leverages principles from cost-aware \emph{multi-fidelity Bayesian optimization} (BO)~\citep{kandasamy2017multi,wu2020practical} to balance the computational cost of running experiments at different scales against the information gained from each experiment. Our method employs a Gaussian Process (GP)~\citep{gp-original-book} as a surrogate model to capture how the LLM loss varies with respect to hyperparameters and scale. Concretely, our contributions are as follows:

\begin{enumerate}
    \item We introduce a new acquisition function, called Power-Law Entropy Search (\textbf{PLES}), that proposes experimentation candidates to efficiently estimate optimal hyperparameter scaling laws. Unlike conventional optimization problems that optimize a single objective function, \textbf{PLES} exploits information from experiments at multiple scales to accurately estimate the entire functional form of a scaling law.

\item We demonstrate that acquisition optimization in \textbf{PLES} can be solved by drawing Thompson samples from the GP, and the sampled solution converges to the true solution with sufficient samples.

\item We empirically demonstrate that, under the same computational budget, \textbf{PLES} produces significantly more accurate optimal hyperparameter scaling laws than conventional baselines in a variety of problems, including hyperparameter optimization for LLM pretraining. In particular, the cumulative computational cost of running small-scale experiments with candidates suggested by \textbf{PLES} is significantly lower than if we had performed hyperparameter optimization directly at the largest scale.
\end{enumerate}

\begin{figure}[h]
    \centering
    \includegraphics[width=\textwidth]{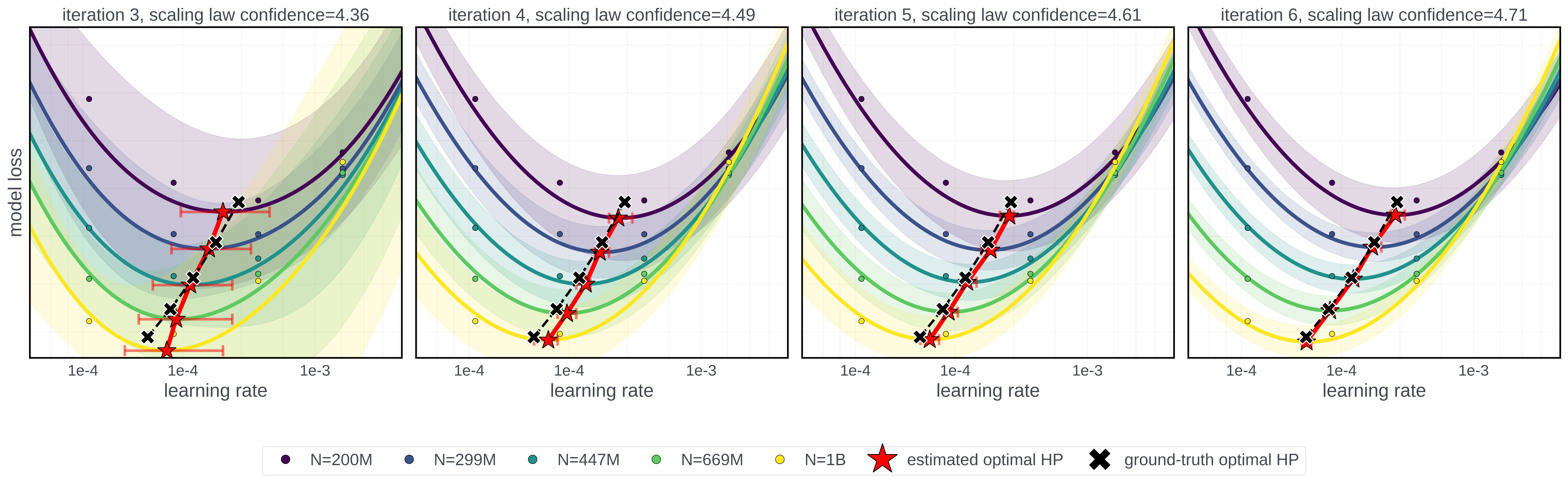}
    \caption{Through small-scale experiments that share information across different model and data scales, \textbf{PLES} produces an optimal hyperparameter scaling law estimate (\textcolor{red}{red line}) that converges to the ground-truth optimal hyperparameter scaling law (black dotted line). In addition, \textbf{PLES} produces uncertainty intervals (red horizontal bars) that indicate our confidence of the scaling law estimate.}
    \label{fig:overview}
\end{figure}


\section{Related Works}

\textbf{LLM Scaling Laws.} Conventional LLM Scaling laws \citep{kaplan2020scaling,hoffmann2022trainingcomputeoptimallargelanguage} describe how LLM validation loss varies with changing model and data scales. Optimal hyperparameter scaling laws \citep{yang2021tuning,mccandlish2018empirical} offer a nuanced distinction from conventional scaling laws because they describe how the \textit{optimal} hyperparameters (conditioned on a given model and data scale) change with increasing model and data scales. Our aim is to learn optimal hyperparameter scaling laws through a series of experimentation at small model and data scales to infer the optimal hyperparameters at larger, production-level model and data scales.

\textbf{Experimental design.} Bayesian Experimental Design (BED) offers a principled statistical framework for selecting experiments that maximize the information gained about a latent quantity of interest, typically quantified via information gain \citep{chaloner1995bayesianDiffEntropy}. Our work leverages some of the same principles found in BED (more details covered in ~\Cref{sec:method}) and uses information gain as a latent quality to drive \textit{where} we perform small-scale experiments. As far as we know, \citep{li2026spend} is the only piece of work that leverages similar principles in designing experiments to learn scaling laws. However, they did not focus on the specific problem of \textit{optimal hyperparameters} scaling across model and data scales, which our paper focuses on. In addition, their approach centers around a Gaussian mixture model formulation in modeling the scaling law parameters while our paper adopts a multi-fidelity BO approach. Last but not least, \cite{li2026spend}'s empirical results are produced from hypothetical experiments from an existing scaling law benchmark, whereas we verified our method on actual LLM training runs (see \Cref{sec:exp}).

\textbf{Gaussian Process (GP)}. A GP \citep{gp-original-book} defines a distribution over functions, $f \sim \mathcal{GP}(\mu, \kappa)$, meaning that any random draw from the process yields a continuous function over the input domain. A GP is fully specified by its \emph{prior} mean function $\mu(x)$ and a covariance kernel function $\kappa(x,x')$ for all $x,x' \in \mathbb{R}^d$, where the kernel characterizes the correlation strength between any two input variables. Conditioned on a set of past observations, $\mathcal{D}_t = \{(x_\tau, y_\tau)\}_{\tau=1}^t$, a GP's posterior belief of any new point $x'$ is a Gaussian distribution, whose mean and variance are:
\begin{align}
\label{gp:posterior}
    \mu_t(x') &\triangleq \kappa_t^{\top}(x')(K_t + \zeta I)^{-1}\bm{y}_t \\
    \sigma_t^2(x') &\triangleq \kappa(x',x')-\kappa_t^{\top}(x')(K_t + \zeta I)^{-1}\kappa_t(x')
\end{align}
where $\kappa_t(x') \triangleq [\kappa(x', x_1), \dots, \kappa(x', x_t)]^{\top}$, $K_t \triangleq [\kappa(x_\tau,x_{\tau'})]_{\tau,\tau' \in \{1,\ldots,t\}}$ is a $t \times t$ covariance matrix, and $\zeta > 0$ is a regularization hyperparameter \citep{bo-kernelized-bandits} and also describes the observation noise under the GP model. A GP is commonly used as a surrogate function in BO algorithms \cite{chen2026duetoptimizingtrainingdata, chen2026chickeneggdilemmacooptimizing,bo-gp-ucb-10}.

\section{Problem setup}\label{sec:setup}
\begin{wrapfigure}{r}{0.42\textwidth}
    \centering
    \includegraphics[width=0.4\textwidth]{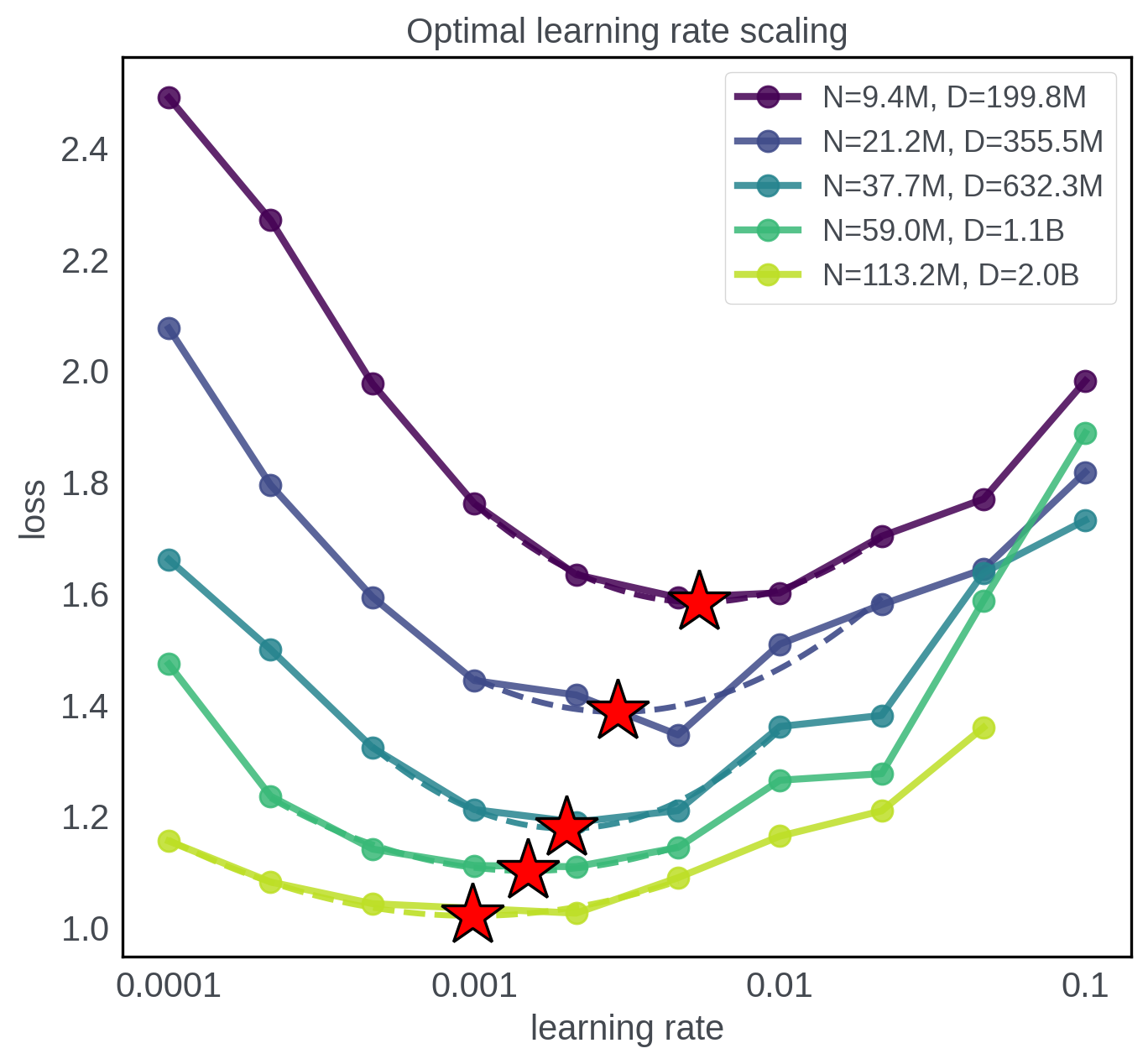}
    \caption{An example of how the optimal learning rate of Llama-3 \citep{touvron2023llamaopenefficientfoundation} changes with increasing model parameters and training data tokens. The change in optimal learning rate (red stars) is described by an optimal hyperparameter scaling law, which we aim to estimate efficiently.}
    \label{fig:example_scaling_law_LR}
\end{wrapfigure}
Let $N \in \mathbb{R}$ denote model size, $D \in \mathbb{R}$ data size, and $\theta \in \mathbb{R}$ a hyperparameter of interest. To ease our exposition, we assume that there is only one hyperparameter, although in ~\Cref{sec:extension-more-hyperparam} we will extend our formulation to more than one hyperparameters (and explain the subtleties involved). Training a model with configuration $(N, D, \theta)$ yields an evaluation loss $\ell(N, D, \theta)$ and consumes some amount of training budget (e.g., GPU hours, FLOPs), often approximated with $6ND$ \citep{kaplan2020scaling}. At each model and data scale, the hyperparameter that minimizes the evaluation loss is referred to as the \textit{optimal hyperparameter}. An example of how the optimal learning rate (red star) scales with increasing model and data sizes is shown in \Cref{fig:example_scaling_law_LR}.

\textbf{Optimal hyperparameter scaling law}. The optimal hyperparameter scaling law describes how the optimal hyperparameter changes with scale. More formally, an optimal hyperparameter scaling law is a predictive function $\Pi$ that maps a set of "scale" parameters to optimal hyperparameters. In our paper, we specifically focus on LLM model scale $N$ and data scale $D$, and $\Pi$ maps these scales to the optimal hyperparameter $\theta^*$ that minimizes the model evaluation loss at that scale. This relationship is described with the following expression: $$\theta^* = \Pi(N, D)$$ where $\Pi: \mathbb{R}^+ \times \mathbb{R}^+ \rightarrow \mathbb{R}$ is a parametric function with learnable coefficients. We are interested in finding the optimal hyperparameter $\tmax$ at the largest target model scale $\Nmax$ and data scale $\Dmax$ which minimizes the evaluation loss, i.e., $\tmax = \argmin_\theta \ell(\Nmax, \Dmax, \theta)$. From hereon, we refer to the largest scale as the \textit{held-out scale}. Assuming the scaling law is known (more specifically, if its coefficients are estimated accurately), the optimal hyperparameter at the held-out scale can be inferred directly from the scaling law: $$\tmax = \Pi(\Nmax, \Dmax).$$

\textbf{Power-law as functional form of scaling law}. In our paper, as with \citep{li2025predictablescaleistep}, we assume that the optimal hyperparameter scaling law $\Pi$ is a power-law function of both model scale $N$ and data scale $D$ (as we show in \Cref{fig:analysis_on_gp_LR,fig:analysis_on_gp_BS}, this assumption is realistic):
  \begin{equation}\label{eqn:power-law}
  \theta^* = c N^{\alpha} D^{\beta}.
  \end{equation}
As such, finding the power-law coefficients is equivalent to finding the optimal hyperparameter scaling law. Our aim is to estimate the power-law coefficients $c,\alpha,\beta$ efficiently through small-scale experimentation.



\section{Method}
\label{sec:method}

We first provide a brief overview of our approach, before covering its details in the next few subsections. Our approach is a two-step process under the Bayesian Optimization (BO) framework where we perform sequential experimentation. At each iteration, (a) we maintain a GP surrogate fitted on previously experimented configurations $(N,D,\theta)$ and observed model loss. We use Thompson samples from the GP to maintain an estimate of the optimal hyperparameter scaling law. (b) Using a new cost-aware acquisition function called \textbf{PLES}, we find the candidate hyperparameter configuration to experiment next that \textit{maximally} reduces the uncertainty of our scaling law estimate. Asymptotically, our estimate converges to the true optimal hyperparameter scaling law.

\subsection{Estimating Power-Law coefficients from a GP}\label{subsec:estimate-pl-from-gp}

Given a fitted Gaussian Process (GP) that takes in input configurations $(N,D,\theta)$, we introduce a procedure to recover the power-law and quantify our uncertainty about its coefficients at any point during experimentation. To achieve this, we rewrite the power-law in \Cref{eqn:power-law} into a linear form:
\begin{equation}
\log \theta^* = \log c + \alpha \log N + \beta \log D.
\end{equation}
We view this as a Bayesian linear regression problem that learns the power-law coefficients $\mathbf{w} = (\log c, \alpha, \beta)^\top$ and their uncertainty. If we are certain about the coefficients, it implies we are certain about the power-law.

\textbf{Finding $\theta^*$ for one particular model $N_i$ and data scale $D_i$}. For a single, fixed model and data scale $(N_i, D_i)$, we perform Thompson sampling \citep{thompson} over the current GP ($\mathcal{GP}$) fitted from observed data to empirically estimate the expectation of the optimal hyperparameter $\theta_i^*$ at scale $i$:
\begin{equation}
    \theta_i^* = \mathbb{E}_{f \sim \mathcal{GP}}[\argmin_{\theta} f(N_i, D_i, \theta)] \approx \frac{1}{K}\sum_{k=1}^{K}\argmin_{\theta} f_k(N_i, D_i, \theta)
    \label{eq:inner-thompson-sample}
\end{equation}
where $f_k$ is one of $K$ Thompson samples from the GP. We also observe that sufficient Thompson samples asymptotically recover the true solution.  We repeat this procedure across $m$ model $[N_1,\cdots,N_m]$ and data $[D_1,\cdots,D_m]$ scales to produce an estimate of optimal hyperparameters across different scales $[\theta_1^*,\cdots,\theta_m^*]$, along with their empirical standard deviations $[\sigma(\theta_1^*),\cdots,\sigma(\theta_m^*)]$.

\textbf{Using Bayesian linear regression to estimate power-law coefficients $\mathbf{w}$}. \label{paragraph:bayesian-lr} We aggregate the evaluations across all $m$ scales to obtain target vector $\mathbf{y}$, the design matrix $\mathbf{X}$, and the diagonal noise covariance matrix $\bm{\Lambda}$:
\begin{equation}
\mathbf{y} = [\log \theta_1^*, \cdots, \log \theta_m]^\top, \quad
\mathbf{X} = 
\begin{bmatrix}
1 & \log N_1 & \log D_1 \\
& \vdots & \\
1 & \log N_m & \log D_m
\end{bmatrix}
, \quad
\bm{\Lambda} = \mathrm{diag}(\sigma^2(\log \theta_1^*), \dots, \sigma^2(\log \theta_m^*)),
\label{eq:BLR-y-x-w}
\end{equation}
where $\sigma(\log \theta_i^*) \approx \sigma(\theta_i^*)/\theta_i^*$\citep{casella2024statistical}.
The power-law is fitted based on $\mathbf{y} = \mathbf{X}\mathbf{w} + \bm{\epsilon}$, where $\bm{\epsilon} \sim \mathcal{N}(\mathbf{0}, \bm{\Lambda})$. Given a Gaussian prior $\mathbf{w} \sim \mathcal{N}(\mathbf{0}, \bm{\Sigma}_0)$ \footnote{We can use a uniform prior if no prior knowledge of the power-law coefficients is known. Otherwise, we can consider using more informative priors.}, our estimate of the power-law coefficients is $p(\mathbf{w} \mid \mathbf{y}, \mathbf{X}) = \mathcal{N}(\bm{\mu}_{\mathbf{w}}, \bm{\Sigma}_{\mathbf{w}})$ , with mean and variance:
\begin{align}
    \bm{\Sigma}_{\mathbf{w}} &= \left( \bm{\Sigma}_0^{-1} + \mathbf{X}^\top \bm{\Lambda}^{-1} \mathbf{X} \right)^{-1} \\
    \bm{\mu}_{\mathbf{w}} &= \bm{\Sigma}_{\mathbf{w}} \mathbf{X}^\top \bm{\Lambda}^{-1} \mathbf{y}
\end{align}
From hereon, we use $\mathbf{\Sigma_w}(\mathcal{GP(D)})$ and $\mathbf{\bm{\mu}_w}(\mathcal{GP(D)})$ to denote the covariance and mean of power-law coefficients estimated from the GP fitted on observed data $\mathcal{D}$.

\textbf{Uncertainty of power-law coefficients}.
The differential entropy $\mathbf{logdet}\bm{\Sigma_w}(\mathcal{GP(D)})$ indicates the amount of uncertainty surrounding the current power-law coefficients estimated from observed data $\mathcal{D}$ \citep{chaloner1995bayesianDiffEntropy}. We use this expression to represent the amount of uncertainty associated with the power-law estimate at any given iteration.

\subsection{Power-Law Entropy Search (PLES)}\label{subsec:ples}

\begin{figure}[h]
    \centering
    \includegraphics[width=0.6\textwidth]{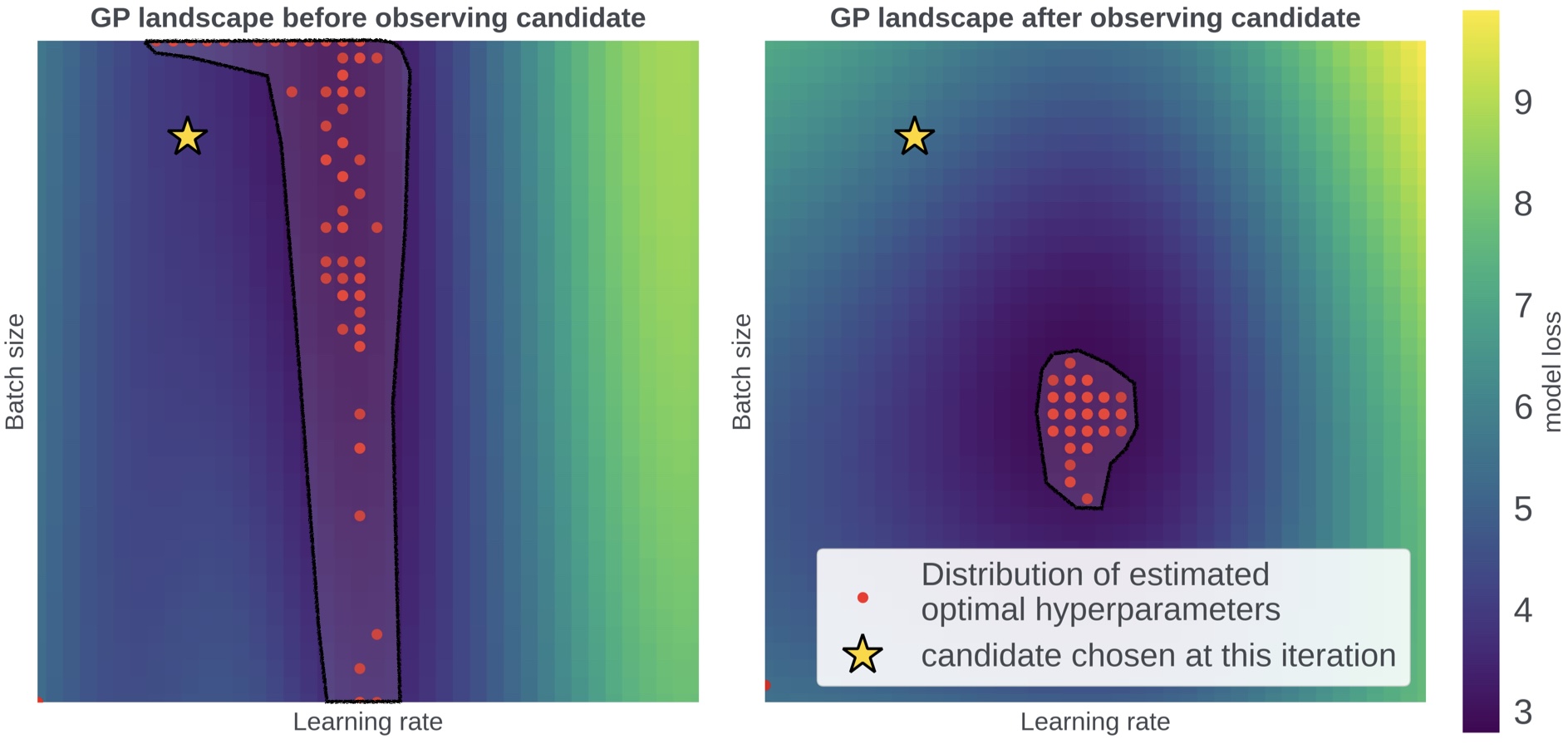}
    \caption{PLES identifies an informative candidate that greatly reduces the empirical uncertainty of the estimated optimal hyperparameters from the GP. This leads to lower variance about the estimated power-law coefficients, which is represented by the expression $\mathbf{logdet\Sigma_w}(\mathcal{GP}(\mathcal{D} \cup x))$ in \Cref{eq:PLES-acquisition}.}
    \label{fig:acquisition_step}
\end{figure}

We introduce a computational cost-aware acquisition function, \textbf{PLES}, that searches for candidates that have the largest uncertainty reduction per unit-cost of the power-law coefficients derived from the procedure in \Cref{subsec:estimate-pl-from-gp}. The overall pseudocode for \textbf{PLES} is provided in \Cref{app:pseudocode-ples}. At iteration $t$, The acquisition function of a candidate hyperparameter and scale $x=(N,D,\theta)$ given historical observed data $\mathcal{D}_t$ is:

\begin{equation}
\label{eq:PLES-acquisition}
    \textbf{PLES}(x) = \argmax_x \frac{\mathbf{logdet \Sigma_w}(\mathcal{GP(D)}) - \mathbf{logdet\Sigma_w}(\mathcal{GP}(\mathcal{D} \cup x))}{(ND)^d},
\end{equation}

where $d$ is a computational cost-cooling factor that discourages experimentation at larger model and data scales \citep{kandasamy2017multi}. If $d=1$ is chosen, it reduces the acquisition function to search for the largest information gain per unit-cost of experimentation. Choosing a larger $d$ favors experimentation at smaller scales, and vice versa. Note that to obtain an updated $\mathcal{GP}(\mathcal{D} \cup x)$, we draw a fantasy observation for $x$ from the current GP. The reduction of the shaded region in \Cref{fig:acquisition_step} describes how an informative candidate reduces the empirical uncertainty of the optimal hyperparameter at a given scale.

\subsection{Extension to more than one hyperparameters}\label{sec:extension-more-hyperparam}

Often, there is more than one hyperparameter of interest when practitioners train LLMs. Our GP modeling process in \Cref{sec:setup} can be extended straightforwardly to these cases, by assuming $\theta$ has a cardinality of more than one. However, the scaling laws of different hyperparameters of LLMs (e.g., learning rate, batch size) are governed by independent power-laws with different coefficients \citep{li2025predictablescaleistep}. To infer the optimal hyperparameters at the largest scale, we will need to maintain an estimate of different, independent power-laws (\Cref{eqn:power-law}) during the BO process. Then, we need to modify our acquisition function in \Cref{eq:PLES-acquisition} to search for candidates that reduce the uncertainty of multiple scaling laws \textit{simultaneously}, by reducing a weighted-sum of differential entropy (mathematical formulation provided in \Cref{app:multi-hp-ples}).


Interestingly, we observed in our experiments that explicitly maintaining multiple scaling laws may not always be necessary. For example, in the case of two hyperparameters (e.g., learning rate and batch size), we find that performing acquisition solely to reduce uncertainty in the scaling law for learning rate can, as a byproduct, also recover an accurate scaling law for batch size. We hypothesize that this occurs because \textbf{PLES}, in identifying configurations that are informative about the optimal learning rate, naturally also reveals regions with the optimal batch size. We analyze these results in greater detail in \Cref{sec:extension-more-than-one}.

\subsection{Stopping criterion}\label{sec:stopping-criteria} A key advantage of maintaining a probabilistic estimate of the power-law coefficients is that it provides a natural stopping criterion for experimentation. At each iteration $t$, the empirical uncertainty of the optimal hyperparameter at the held-out scale is: $ \mathbf{x}_{\text{max}}^\top \bm{\Sigma_w} \mathbf{x}_{\text{max}}$, where $\mathbf{x}_{\text{max}} = (1, \log N_{\text{max}}, \log D_{\text{max}})^\top$ (see \Cref{paragraph:bayesian-lr}). Essentially, this is the uncertainty of the optimal hyperparameters (in the power-law) at the held-out scale. When this value is small, it suggests that our power-law estimate has become accurate. This is shown in column 3 and 4 of \Cref{fig:analysis_on_gp_LR}, where the uncertainty at the held-out scale is small and further experimentation does not improve the estimated power-law significantly.

\section{Experiments}
\label{sec:exp}

\subsection{Setup}
\label{sec:exp:setup}

We evaluate our method on a few settings, including \textit{real LLM training runs}. \textbf{First}, we consider a synthetic function that describes how model loss varies with respect to varying learning rate, batch size, model scale $N$ and data scale $D$. The function's formula is provided in \Cref{app:synthetic-details}. \textbf{Second}, we retrieved past LLM training runs from \citep{li2025predictablescaleistep, lin2026languagemodelsdiscoverscaling}. The training runs contain a grid search of how model loss varies w.r.t. learning rate, batch size, model and data scales. We fit a GP to the results of these training runs, use the GP as an oracle function and apply our method on a continuous set of configuration choices at every acquisition step. \Cref{app:gp-fit} shows that the GP has a good fit over the past LLM training run data, serving as a good surrogate ground-truth. \textbf{Third}, we evaluate our method on \textit{real LLM training runs}, whose training setup we cover next.

\textbf{Setup for real LLM training runs.} For simplicity, we use one dimension of scaling and only vary the model size $N$. For each model size chosen, we use a data size of 20 Tokens-Per-Parameter (TPP), which follows the Chinchilla Law \citep{hoffmann2022trainingcomputeoptimallargelanguage}. We train a Llama-8B model. To adjust the model size, we vary the number of layers and hidden layer dimensions in the intermediate layers simultaneously, while keeping the model's aspect ratio roughly the same. The range of configurations used is as follows: $N \in [9.4M, 600M], D \in [199.8M, 12B], \textrm{LR} \in [0.0001,0.1].$ For every configuration, we pre-train the LLM using a subset of the Llama training data \citep{touvron2023llamaopenefficientfoundation} for one epoch and record the final validation loss on a held-out set from the training data. We use a batch size of 2048 and the sequence length of each batch is 2048 tokens.

\textbf{Evaluation methods.} We evaluate the effectiveness of our method by evaluating how good the optimal hyperparameter at the held-out scale is. There are two ways to do so. For the \textbf{synthetic} and \textbf{surrogate} experiments, we know the ground-truth optimal $\theta$ at the held-out scale. For the \textbf{Real LLM} experiment, we performed grid search and interpolation at the held-out scale to empirically estimate the optimal $\theta$. We then measure the percentage error of the optimal $\theta$ inferred from the estimated power-law at the held-out scale. The held-out scale is indicated by the figure subtitles in \Cref{fig:results_overall}. The second evaluation metric is to evaluate the model loss of the estimated optimal $\theta$ at the held-out scale at the end of every acquisition step.

\subsection{Baselines}
\label{sec:baselines}

We compare our method against three baselines. All baselines operate under the same total
compute budget (measured as the cumulative $ND$ spent across all evaluations)
and, where applicable, share the same Sobol-initialized starting configurations to ensure a fair comparison. For every method, we repeat our experiments with 10 seeds, which influence the initial starting configurations and also the randomness involved in every method.

\textbf{Grid search.} This is the conventional method used to fit scaling laws in
prior work \citep{hoffmann2022trainingcomputeoptimallargelanguage,
li2025predictablescaleistep}. For each model/data scale $(N, D)$, we evaluate the loss on a dense grid over the hyperparameters
$\theta$, take the per-scale grid minimizer as the estimated
optimal hyperparameter $\theta^\star(N, D)$, and then fit the scaling law. The budget is allocated such that every scale contains approximately the same number of training runs. There are many ways to allocate compute resources for grid search, but we find this approach more effective than others. We provide a more thorough analysis of budget allocation of grid search in \Cref{app:budget-allocation-grid-search}.

\textbf{Ladder BO.} This baseline performs independent 
Bayesian optimization with the Expected Improvement \citep{ament2023unexpected} to estimate the optimal hyperparameter $\theta$ at each scale. To achieve this, independent GPs are fitted to the results at each scale, and computational resources are allocated such that we have an equal amount of training runs at each scale. The optimal hyperparameter estimated at each scale is then used to fit the scaling law, allowing us to estimate the optimal hyperparameter at the held-out scale.

\textbf{Sobol.} This baseline draws
configurations $(\theta, N, D)$ from a quasi-random, space-filling Sobol sequence
until the budget is exhausted, then fits a GP over the hyperparameter $\theta$
and scale $(N, D)$. The scaling law is estimated from the random configurations using the same procedure as in \Cref{subsec:estimate-pl-from-gp}.

\subsection{Main Results}
\label{sec:main-results}

\begin{figure}[h]
    \centering
    \includegraphics[width=1\textwidth]{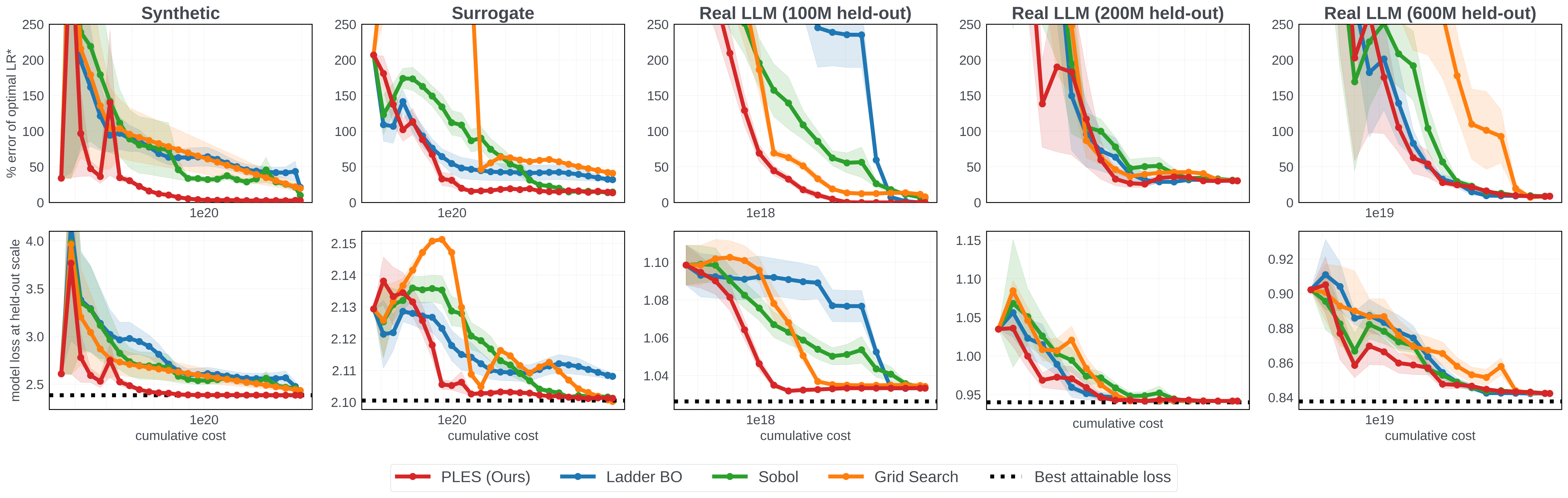}
    
    \caption{Results on three settings: a \textbf{Synthetic function}, a \textbf{Surrogate} GP model that is fitted on real LLM training runs in \citep{li2025predictablescaleistep}, and \textbf{Real LLM} training runs on the Llama-3 \citep{touvron2023llamaopenefficientfoundation} model. The first row shows how accurate the optimal hyperparameter estimated by each method is w.r.t. increasing experimentation computational cost. The second row shows the model loss attained at the estimated optimal hyperparameter at the held-out scale.}
    \label{fig:results_overall}
\end{figure}

Our results in \Cref{fig:results_overall} show that \textbf{PLES} (\textcolor{red}{red}) consistently achieves a smaller percentage error in the estimated optimal hyperparameter than other baselines in the held-out scale. In addition, the optimal hyperparameter estimated by \textbf{PLES} converges much more quickly than other methods, reaching the best attainable model loss at the held-out scale in \textit{less than one-tenth} of the computational budget as compared to canonical baselines.

\subsection{Analysis of scaling law uncertainty with increasing computational cost}\label{sec:analysis-on-fitted-gp}

\begin{figure}[h]
    \centering
    \includegraphics[width=1.0\textwidth]{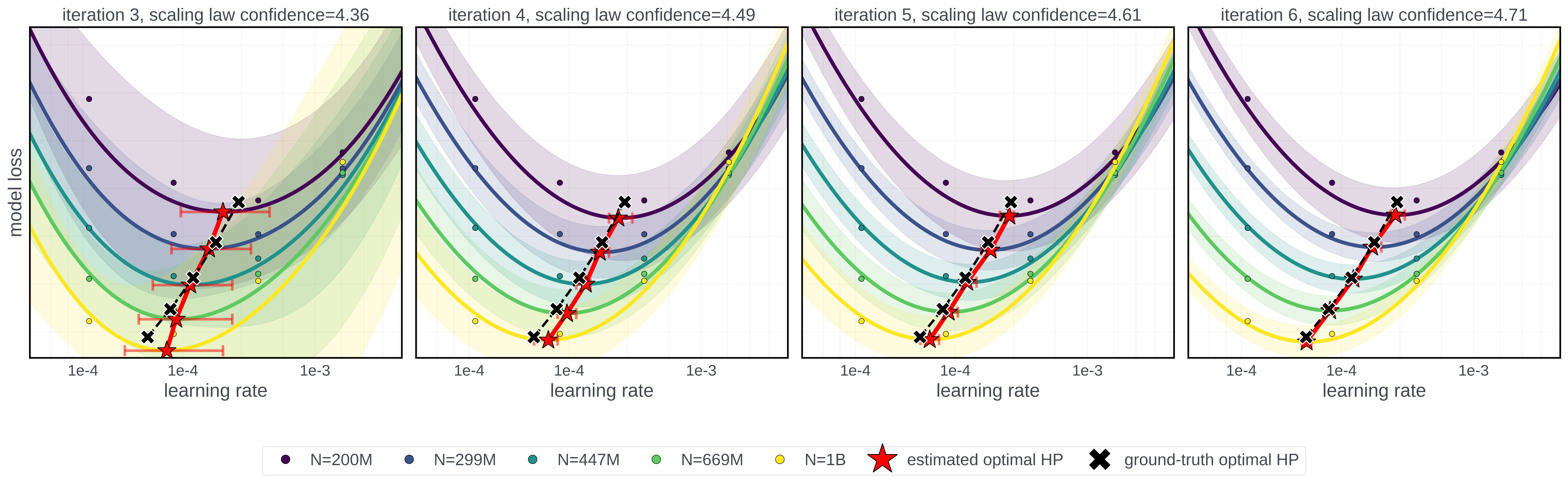}
    \caption{The fitted GP and estimated optimal learning rate scaling law at each iteration of the \textbf{surrogate} experiment. The variance of the GP reduces and the estimated scaling law converges to the ground-truth scaling law with more iteration. At iteration 6 (3rd column), the uncertainty at the held-out scale is sufficiently small, serving as a stopping criterion for the experiment. Further experimentation at the 4th column does not improve the estimated scaling law significantly.}
    \label{fig:analysis_on_gp_LR}
\end{figure}


\begin{figure}[h]
    \centering
    \includegraphics[width=1.0\textwidth]{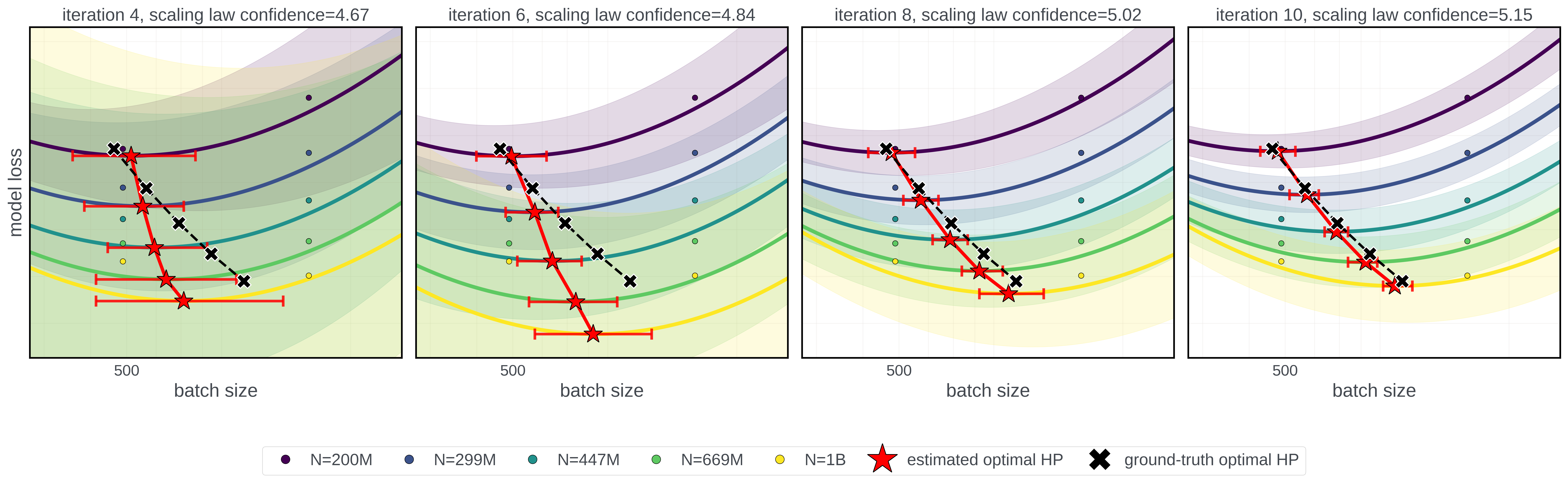}
    \caption{Although \textbf{PLES} does not explicitly model the optimal batch size scaling law for the \textbf{surrogate} experiment, it still converges to the ground-truth optimum.}
    \label{fig:analysis_on_gp_BS}
\end{figure}

In this section, we analyze how the confidence of our estimated scaling law changes as we incur more experimentation budget in the \textbf{surrogate} experiment. The size of the horizontal standard deviation bar in \Cref{fig:analysis_on_gp_LR} represents our uncertainty of the learning rate scaling law fit $\bm{\Lambda}$ (see \Cref{eq:BLR-y-x-w}) at each iteration. We observe that the uncertainty decreases with more experimentation budget. In addition, the size of the horizontal standard deviation at the largest held-out scale (in this case, at N=1B) can also be used as a stopping criterion (see \Cref{sec:stopping-criteria}) in PLES. At iteration 6 (3rd column), the uncertainty of the estimated scaling at the held-out scale is small, suggesting that our scaling law estimate has (rightfully so) already converged to the ground-truth scaling law. As expected, performing experiments for more iterations (from 3rd to 4th column) does not improve the scaling law estimate significantly.

\subsection{Results on more than one hyperparameters (learning rate and batch size)}\label{sec:extension-more-than-one}

In \Cref{sec:extension-more-hyperparam}, we mentioned that PLES can recover multiple optimal hyperparameter scaling laws even when it is designed to minimize the uncertainty about one scaling law. In the \textbf{surrogate} experiment, our GP takes in both learning rate and batch size but we designed \textbf{PLES} to specifically acquire candidates that reduce the uncertainty of the learning rate scaling law. Despite so, the result in \Cref{fig:analysis_on_gp_BS} shows that PLES surprisingly recovers the batch size scaling law despite only being designed to estimate the learning rate scaling law. One possible explanation for this phenomenon is that PLES uncovers regions that contain the optima of multiple hyperparameters simultaneously; this is seen in \Cref{fig:acquisition_step}, where an acquired candidate (yellow star) that reduces the uncertainty of one hyperparameter (e.g., learning rate) also reduces the uncertainty of another hyperparameter (e.g., batch size).

\subsection{Analysis of the scale of candidates acquired by \textbf{PLES}}
\label{sec:analysis-candidate-scale}

\begin{wrapfigure}{r}{0.42\textwidth}
    \centering
    \vspace{-5mm}
    \includegraphics[width=0.4\textwidth]{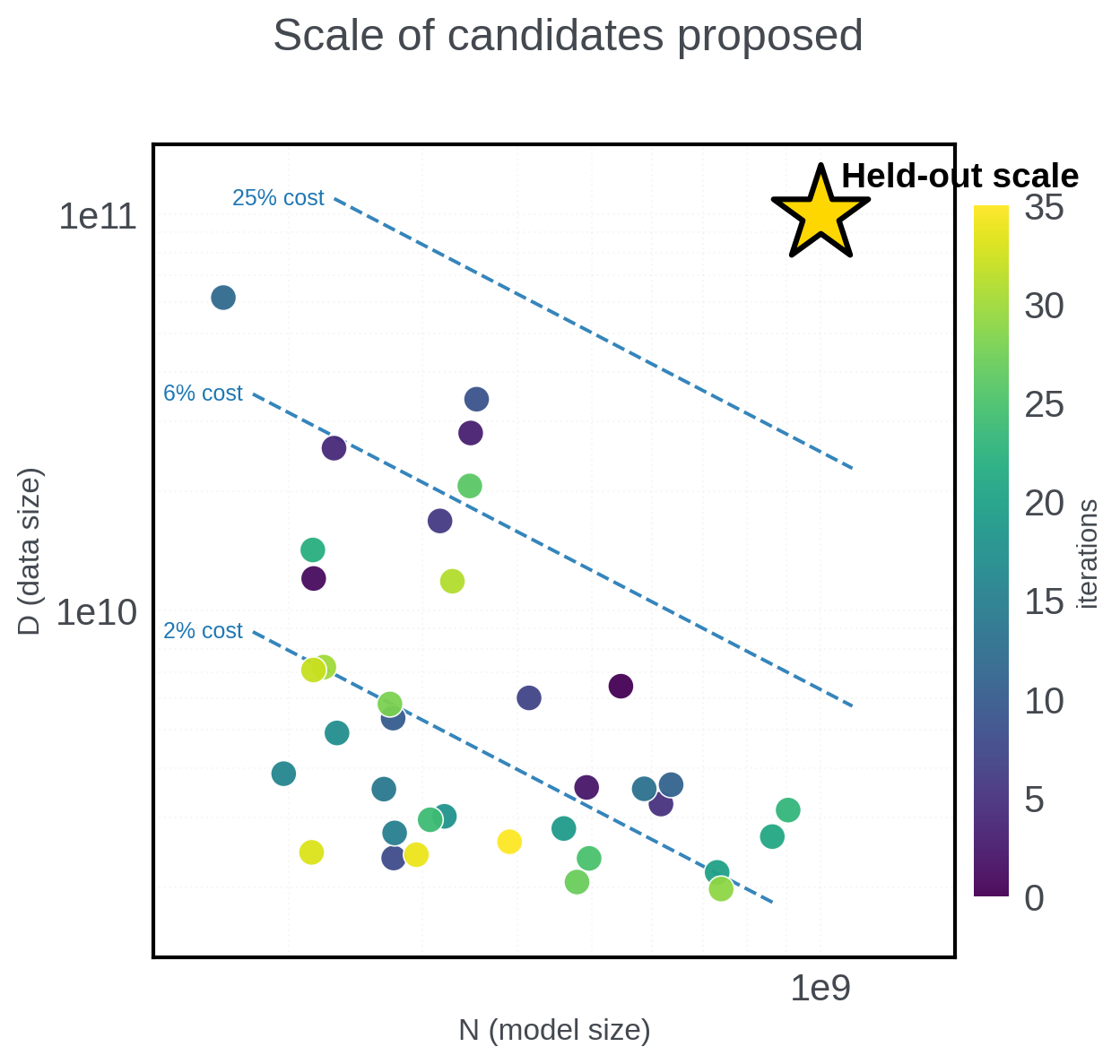}
    \caption{The candidate scales that \textbf{PLES} proposes throughout the course of learning the optimal hyperparameter scaling law. Most candidate scales are less than $2\%$ of the computational cost of the held-out scale.}
    \label{fig:candidate-scales}
\end{wrapfigure}
\Cref{fig:candidate-scales} shows that most of the candidate scales proposed by acquisition function \textbf{PLES} are concentrated at small scales that are less than $6\%$ of the computational cost of running the experiments at the expensive held-out scale. This is because \textbf{PLES} is computational cost-aware and uses a cost-cooling factor in \Cref{eq:PLES-acquisition} to favor candidates at cheaper scales. As a result, the computational cost of running all the experiments with the candidates suggested by \textbf{PLES} is significantly smaller than the computational cost of performing hyperparameter optimization at the held-out scale directly.

\subsection{Computational complexity}\label{sec:computational-complexity}

\textbf{PLES} allows us to recover the optimal hyperparameter scaling law in an efficient amount of computational budget. Some computational complexity is necessary to evaluate the acquisition value of each candidate configuration and solve the optimization problem in \Cref{eq:PLES-acquisition}. The per-iteration complexity is $\mathcal{O}(C \cdot K \cdot Q)$, where $C$ is the number of candidate configurations we consider for the optimization problem, $K$ is the number of Thompson samples used to estimate $\theta^*$ in \Cref{subsec:estimate-pl-from-gp}, and $Q$ is the computational cost of solving the inner $\arg\min_\theta$ problem (\Cref{eq:inner-thompson-sample}) for each sample. Leveraging batched pathwise sampling and vectorized operations in BoTorch \citep{balandat2020botorch}, the Thompson sampling loop and inner minimizations can be fully parallelized across $K$ and $C$. Hence, the computational overhead in \textbf{PLES} is negligible compared to the reduction in the amount of model training computational cost saved.

\section{Limitations}
\label{sec:limitation}

Our method assumes that the optimal hyperparameter scaling law follows a known functional form (\Cref{eqn:power-law}). Although this assumption is well-supported empirically for the hyperparameters and architectures studied in this paper (\Cref{fig:analysis_on_gp_LR,fig:analysis_on_gp_BS}) and in prior work~\citep{li2025predictablescaleistep}, it may not hold universally. It is possible that the trend of optimal hyperparameters deviates from a simple power-law. In such cases, the Bayesian linear regression step in \Cref{subsec:estimate-pl-from-gp} would produce biased coefficient estimates and degrade the extrapolation accuracy at the held-out scale. That said, extending PLES to accommodate more flexible functional representations while retaining tractable uncertainty quantification is a promising direction for future work.

\section{Conclusion}
\label{sec:conclusion}

We introduced Power-Law Entropy Search (\textbf{PLES}), a cost-aware acquisition function that efficiently estimates optimal hyperparameter scaling laws for large language model training through efficient experimentation.

Our experiments across synthetic benchmarks, surrogate models fitted to real LLM training data, and actual LLM pre-training runs demonstrate that \textbf{PLES} converges to accurate optimal hyperparameter scaling laws using less than one-tenth of the computational budget required by conventional grid search and other baselines. Furthermore, \textbf{PLES} provides a natural stopping criterion based on the posterior uncertainty at the target scale, allowing practitioners to terminate experimentation once sufficient confidence is achieved. We also showed that \textbf{PLES} can recover scaling laws for multiple hyperparameters simultaneously, even when the acquisition function is designed to reduce uncertainty for only one.

We believe \textbf{PLES} offers a practical and principled framework for scaling law estimation that can substantially reduce the computational cost of developing future large language models. As model architectures and training recipes continue to evolve---requiring repeated re-estimation of scaling laws---methods like \textbf{PLES} that minimize the experimental burden become increasingly valuable.

\clearpage
\newpage
\bibliographystyle{assets/plainnat}
\bibliography{paper}

\clearpage
\newpage
\beginappendix

\section{Pseudocode for \textbf{PLES}}\label{app:pseudocode-ples}

\begin{algorithm}[h]
\begin{tcolorbox}[
  colback=gray!15,        
  colframe=gray!50,
  boxrule=0.6pt, arc=3pt,
  left=4pt, right=4pt, top=4pt, bottom=4pt
]
\SetAlgoLined
\DontPrintSemicolon
\KwIn{Current $\mathcal{GP}$ model fitted on observations prior to iteration $t$; a set of candidate configurations each with model scale $N$, data scale $D$, cost-cooling factor $d$}
\KwOut{Acquisition value for each candidate.}
\BlankLine
Compute current power-law uncertainty $= \mathbf{logdet}\bm{\Sigma_w}(\mathcal{GP}(\mathcal{D}))$\;
\ForEach{candidate}{
  1. Draw a fantasy observation and update GP model: $\mathcal{GP}(\mathcal{D} \cup x)$.\;
  2. Compute updated power-law uncertainty $\mathbf{logdet}\bm{\Sigma_w}(\mathcal{GP}(\mathcal{D} \cup x))$ (\Cref{subsec:estimate-pl-from-gp}).\;
  3. Acquisition value of candidate: $\frac{\mathbf{logdet \Sigma_w}(\mathcal{GP(D)}) - \mathbf{logdet\Sigma_w}(\mathcal{GP}(\mathcal{D} \cup x))}{(ND)^d}$.\;
}
\end{tcolorbox}
\caption{Pseudo code of Power-Law Entropy Search (\textbf{PLES}) at iteration $t$ that computes the acquisition value of each candidate. The candidate with the highest acquisition value is selected.}
\label{alg:ples}
\end{algorithm}

\section{PLES for Multiple Hyperparameters} \label{app:multi-hp-ples} Without loss of generality, we illustrate the extension of PLES to two hyperparameters $\theta^{(1)}$ and $\theta^{(2)}$ (e.g., learning rate and batch size), each with its own independent power-law: \begin{align} \log \theta^{(1)*} &= \log c_1 + \alpha_1 \log N + \beta_1 \log D, \\ \log \theta^{(2)*} &= \log c_2 + \alpha_2 \log N + \beta_2 \log D. \end{align} Following the procedure in \Cref{subsec:estimate-pl-from-gp}, we estimate each set of power-law coefficients independently: \begin{equation} \mathbf{w}_1 = (\log c_1,\; \alpha_1,\; \beta_1)^\top, \quad \mathbf{w}_2 = (\log c_2,\; \alpha_2,\; \beta_2)^\top, \end{equation} each with its own Bayesian linear regression posterior: \begin{equation} p(\mathbf{w}_j \mid \mathcal{D}) = \mathcal{N}(\bm{\mu}_{\mathbf{w}_j},\; \bm{\Sigma}_{\mathbf{w}_j}), \quad j \in \{1, 2\}. \end{equation} These two regression problems share the same design matrix $\mathbf{X}$ (since the $(N_i, D_i)$ scales are common) but have different target vectors $\mathbf{y}_1, \mathbf{y}_2$ and noise matrices $\bm{\Lambda}_1, \bm{\Lambda}_2$, estimated in the same manner as \Cref{subsec:estimate-pl-from-gp}. The modified \textbf{PLES} acquisition function searches for candidates to reduce the uncertainty of both scaling laws \textit{simultaneously} by maximizing the sum of entropy reductions: \begin{equation} \textbf{PLES}(x) = \argmax_x \frac{\ \,\Delta_1(x) \;+\; \,\Delta_2(x)}{(ND)^d}, \end{equation} where \begin{equation} \Delta_j(x) = \textbf{logdet}\,\bm{\Sigma}_{\mathbf{w}_j}(\mathcal{GP}(\mathcal{D})) \;-\; \textbf{logdet}\,\bm{\Sigma}_{\mathbf{w}_j}(\mathcal{GP}(\mathcal{D} \cup x)) \end{equation} is the reduction in differential entropy of the $j$-th scaling law's coefficients after observing candidate $x$.

\section{Details of Synthetic Function}
\label{app:synthetic-details}

To evaluate our method in a controlled setting with a known ground-truth optimum, we construct a four-dimensional synthetic benchmark that mimics the loss landscape of neural language-model training. The function takes two optimization variables — the learning rate $\eta$ and the batch size $B$ (measured in tokens) — together with two fidelity variables, the model size $N$ (parameters) and the data size $D$ (tokens). All four inputs are parameterized in $\log_{10}$ space, and the computational cost of an evaluation is defined as the compute proxy $\text{cost}(N,D)=N\cdot D$, so that low-fidelity (small-$N$, small-$D$) queries are cheap and high-fidelity queries are expensive.

The benchmark is built so that the loss-minimizing hyperparameters follow prescribed scaling laws. The optimal learning rate obeys a joint power law in model and data size, $\eta^{}(N,D)=A,N^{b}D^{c}$ with $(A,b,c)=(0.1896,,-0.734,,0.342)$, so that the optimal learning rate decreases with model size and increases mildly with data; the optimal batch size follows a power law in the data size alone, $B^{}(D)=G,D^{d}$ with $(G,d)=(14.9624,,0.5)$. Around this optimum the loss is modeled as a smooth quadratic "bowl" in log-space, $$ \mathcal{L}(\eta,B,N,D)=L_{\text{floor}}(N,D)+\alpha_\eta\big(\log\eta-\log\eta^{}\big)^{2}+\alpha_B\big(\log B-\log B^{}\big)^{2}, $$ with curvature coefficients $\alpha_\eta=0.40$ and $\alpha_B=0.15$ (deviations measured in natural log; the optional $\eta$–$B$ cross-term is set to zero, giving an axis-aligned bowl). The additive term $L_{\text{floor}}$ is a Chinchilla-style irreducible loss that improves with scale, $$ L_{\text{floor}}(N,D)=L_\infty+\frac{A_N}{N^{\beta_N}}+\frac{A_D}{D^{\beta_D}},\qquad L_\infty=1.69,;(A_N,\beta_N)=(406.4,0.34),;(A_D,\beta_D)=(410.7,0.28). $$ 

\section{Using GP as Oracle Ground-truth}
\label{app:gp-fit}

\begin{figure}[h]
    \centering
    \includegraphics[width=1.0\textwidth]{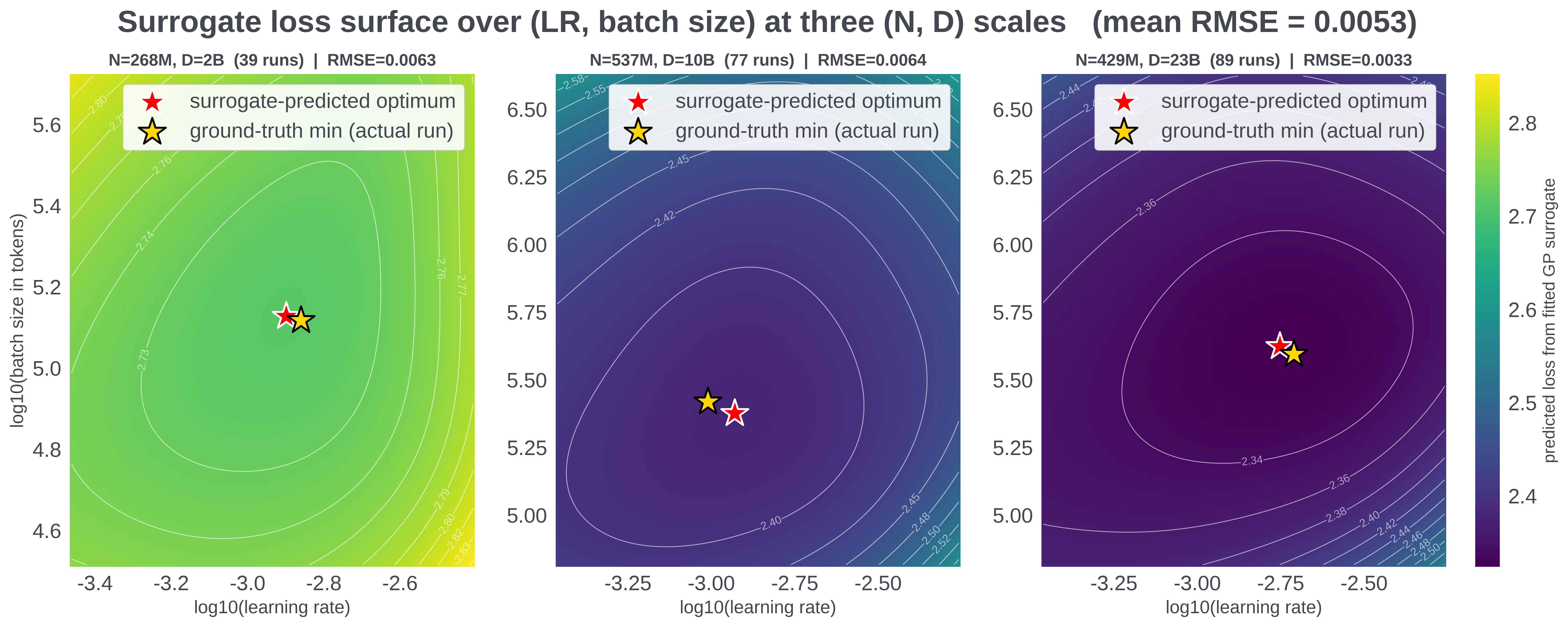}
    \caption{The surrogate loss surface shows that the GP serves as a fit over the data from training runs in \citep{li2025predictablescaleistep}. This implies it can be used as an accurate ground-truth function to validate the \textbf{PLES} algorithm in our \textbf{Surrogate} experiments.}
    \label{fig:GP_fitted}
\end{figure}

In our \textbf{surrogate} experiments, we fitted a surrogate GP over the data from training runs in \citep{li2025predictablescaleistep} and use it as the ground-truth to validate our algorithm, \textbf{PLES}. \Cref{fig:GP_fitted} shows that the surrogate GP is well-fitted over the data from the training runs. Specifically, the optimal learning rate and batch size at each model and data scale from the training runs (yellow star) are close to the predicted optimal hyperparameters from the GP (red star). In addition, the RMSE are all smaller than 0.01, suggesting that the GP surrogate serves as a good fit to represent the data from real training runs.

\section{Budget Allocation for Grid Search}\label{app:budget-allocation-grid-search}

\begin{figure}[h]
    \centering
    \includegraphics[width=1.0\textwidth]{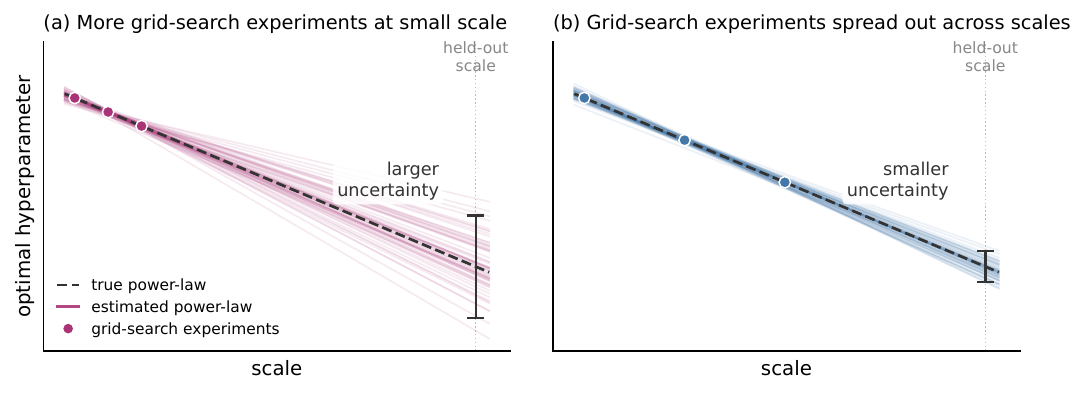}
    \caption{In grid search, we allocate budget uniformly across all scales because it is a stronger baseline than allocating budget focused at a small scale.}
    \label{fig:budget-allocation-grid-search}
\end{figure}

When fitting an optimal hyperparameter scaling law using grid search, a natural instinct is to spend the entire tuning budget on small models, since they are cheap and one can afford many runs to pin down the optimum at small scale precisely. This turns out to be a poor use of compute: predicting the best hyperparameter at a much larger scale is similar to extrapolation in a regression problem, and spreading the budget across more scales is more effective. Points crowded together at small scales pin down the line's slope poorly, and even a tiny slope error would propagate to larger uncertainties at larger scales. Spreading experiments across a wide range of scales gives a better estimate, even though each individual measurement is coarser (Figure~\ref{fig:budget-allocation-grid-search}). Therefore, in our grid search baseline in \Cref{sec:baselines}, we chose to allocate experimental runs evenly across all scales to make it a stronger baseline.

\end{document}